# A Reinforcement Learning-Based Task Mapping Method to Improve the Reliability of Clustered Manycores

Fatemeh Hossein-Khani, Omid Akbari

*Abstract*— The increasing scale of manycore systems poses significant challenges in managing reliability while meeting performance demands. Simultaneously, these systems become more susceptible to different aging mechanisms such as negative-bias temperature instability (NBTI), hot carrier injection (HCI), and thermal cycling (TC), as well as the electromigration (EM) phenomenon. In this paper, we propose a reinforcement learning (RL)-based task mapping method to improve the reliability of manycore systems considering the aforementioned aging mechanisms, which consists of three steps including bin packing, task-to-bin mapping, and task-to-core mapping. In the initial step, a density-based spatial application with noise (DBSCAN) clustering method is employed to compose some clusters (bins) based on the cores' temperature. Then, the Q-learning algorithm is used for the two latter steps, to map the arrived task on a core such that the minimum thermal variation is occurred among all the bins. Compared to the state-of-the-art works, the proposed method is performed during runtime without requiring any parameter to be calculated offline. The effectiveness of the proposed technique is evaluated on 16, 32, and 64 cores systems using SPLASH2 and PARSEC benchmark suite applications. The results demonstrate up to 27% increase in the mean time to failure (MTTF) compared to the state-of-the-art task mapping techniques.

## I. INTRODUCTION

Advancements in technology over the past decade have significantly increased the number of cores on a chip, greatly simplifying the design of cores in both industry and academia [1]. This progress, however, has brought forth several design challenges, with a primary focus on reducing operating voltage, where with the voltage scaling, manycore systems are getting more sensitive to aging mechanisms, including negative bias temperature instability (NBTI), electromigration (EM), and thermal cycling (TC). Furthermore, the technology scaling has resulted in higher power density and chip temperature, accelerating the aging process that may compromise the system's reliability [2]. Also, the manufacturing-induced process variations (PVs) further impact the lifetime of cores [3].

Existing researches have primarily concentrated on improving the health of individual cores [2], minimizing mean time to failure (MTTF) [4], and enhancing the average MTTF of all cores in a manycore system [5], where architecture-level approaches have been extensively explored to achieve these improvements. Notably, dynamic voltage and frequency scaling (DVFS) [6][7], dynamic thermal management (DTM) [8][9], and dynamic reliability management (DRM) [10] have been investigated. Moreover, as task-to-core mapping is one of the most crucial issues in manycores [11], system-level approaches provide dynamic opportunities by controlling task-to-core mappings and per-core operation frequencies to address aging and reliability management challenges [2][3][5]. Despite the considerable attention given to mechanisms like NBTI or EM [12]-[14], thermal cycling has received relatively less focus [15], which is a critical reliability concern. However, thermal cycling is not solely dependent on temperature levels; it also incorporates the amplitude and frequency of temperature variations. Thus, addressing thermal cycling is crucial for effective reliability management [15].

In our previous work [16], we proposed a two-level thermal cycling-aware task mapping technique, in which, at first, cores are packed into bins based on the cores' temperatures, and then, the arrived task is mapped into the appropriate core inside of a given bin. However, in this method, applications are performed offline to extract the required parameters (e.g., temperatures of the tasks) for the mapping approach. Then, during the runtime when applications are executing, the offline calculated task temperatures are employed to compose the bins. To remove this restriction, and motivated by the aforementioned challenges of the aging phenomena, this paper proposes a reinforcement learning (RL)-based task mapping technique that improves the average mean time to failure of all cores by considering the thermal exchange between adjacent cores. This technique employs a mixed-grained task mapping approach in two levels: task-to-bin and task-to-core mapping steps. Initially, based on core temperatures, the cores are organized into bins using the density-based spatial clustering of applications with noise (DBSCAN) algorithm. Afterward, in the task-to-bin mapping step, a bin is selected for mapping the arrived task such that imposes minimum temperature variation among the cores and bins. Then, in the task-to-core step, the task is assigned to a core inside the selected bin such that the power and performance requirements are met. Compared to the [16], in this work, both task mapping levels, task-to-bin and task-to-core mapping level, are performed leveraging the Q-learning algorithm, which removes the need for offline execution of applications, as well as achieving higher MTTF improvement. Figure 1 illustrates the high-level overview of our technique.

F. Hossein-Khani and O. Akbari (*Corres. Author*) are with the Department of Electrical and Computer Engineering, Tarbiat Modares University, Tehran, Iran (e-mails: f.hosseinkhani@modares.ac.ir, o.akbari@modares.ac.ir).

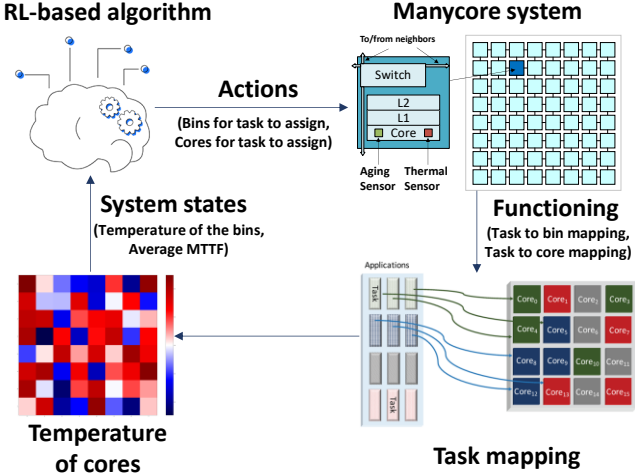

Figure 1. Overview of our proposed technique.

The rest of this paper is organized as follows. In Section II, state-of-the-art works are reviewed. The process variation and reliability models, as well as the employed RL method, are studied in Section III. The proposed task mapping algorithm is introduced in Section IV and the results for the efficacy of the proposed method are discussed in Section V. Finally, this paper is concluded in Section VI.

## II. RELATED WORKS

Reliability is a critical factor that significantly influences the overall performance and functionality of manycore systems. In recent years, numerous approaches have been proposed to address reliability management, taking into account factors such as power consumption, temperature, and aging. These methods often utilize greedy algorithms or leverage the capabilities of machine learning techniques. In this section, we study the state-of-the-art works improving the reliability of manycore systems, and specifically, those works that employed the RL-based methods for manycore systems.

In [2], a reinforcement learning-based approach called LifeGuard was proposed to address process variation and aging challenges of manycores, and especially, to reduce the rapid aging of cores. In [5], authors proposed a dynamic hierarchical mapping approach called HiMap, which aims to maximize the reliability of manycore systems while considering performance, power, and temperature constraints. The approach involves cluster-based mapping at the first level and strategically placing threads for uniform aging within the cluster at the second level. They also leverage dark cores for thermal mitigation by interspersing them within the cluster. Based on HiMap, a hierarchical mapping solution involving voltage and frequency (VF) selection to enhance the lifetime reliability of processors was proposed in [3]. A power management method for manycore systems based on conscious reinforcement learning was proposed in [7]. The primary objective of this method is to create a tradeoff between functional and thermal reliability, while simultaneously achieving power savings and maintaining performance levels. This approach incorporates a reinforcement learner, which takes into account power reduction and the impact of dynamic voltage and frequency scaling (DVFS) on thermal reliability. By estimating power consumption and assessing reliability, the reinforcement learning agent determines the appropriate voltage-frequency level to optimize system performance while ensuring reliability is upheld.

In [8], dark cores were leveraged to mitigate the aging in a systems-on-chip (SoC). This method investigates the interplay between dark cores, temperature, and process variations, and demonstrates how these factors can be synergistically utilized to effectively slow down chip aging and maintain high performance throughout the system's lifetime. To achieve this, the proposed method, named Hayat, selects a subset of cores that meet the performance requirements for concurrent execution of multi-threaded programs while minimizing the overall chip aging.

Considering the challenges posed by temperature variations and thermal cycling, [17] presents a multi-level thermal stress-aware power and temperature management method. Using the DVFS technique, this method mitigates high spatial and temporal thermal variations within a multiprocessor SoC. A method for mapping and scheduling real-time intermittent tasks in multi-core embedded systems, while achieving a certain level of reliability considering the thermal design power (TDP) was proposed in [12], called ReMap. This method uses task duplication to achieve the desired reliability for the target system. In [13], a reliability improvement technique for manycore systems utilizing task duplication to ensure that the power consumption of cores remains below the thermal safe power (TSP) threshold, was proposed.

As discussed, only a few works have focused on mitigating the thermal cycling effects. However, in this work, we propose a RL-based task mapping technique for manycores to improve the system reliability, while considering the heat exchanges between adjacent cores, as well as the effects of process variations. Also, unlike prior state-of-the-art work of [16], in this work, the main parts of the proposed task mapping, including the task-to-bin and task-to-core mapping are performed leveraging the RL algorithm, and thus, there is no need for any offline execution of applications that can gain higher reliability improvement.

## III. BACKGROUND

In this section, an overview of the process variation and reliability models, as well as the RL preliminaries required for the proposed method are presented.

### A. Process variation

In the context of manycore systems, process variation can introduce variations in core characteristics, leading to changes in safe operating frequencies and leakage power [2]. To model the PV in a manycore, a grid-based representation with the dimensions of $P \times Q$ is employed for the chip surface [4]. Within this grid, the process variation information at each grid cell is represented by $p_{(x,y)}$, where $(x,y)$ represents the position of a core. Considering this information, the physical parameters of the cores are given by [3]:

$$W_{x,y} = \kappa_1 p_{x,y} \quad (1)$$
$$H_{x,y} = \kappa_2 p_{x,y} \quad (2)$$
$$Res_{x,y} = \gamma p_{x,y} \quad (3)$$

> REPLACE THIS LINE WITH YOUR MANUSCRIPT ID NUMBER (DOUBLE-CLICK HERE TO EDIT) <   3where $W_{x,y}$, $H_{x,y}$, and $Res_{x,y}$ represent the wire width, wire height, and the power grid resistance at the cell *(x,y)*, respectively. Also, $\kappa_1$, $\kappa_2$, and $\gamma$ are technology-specific constants. Moreover, the maximum frequency of a core is determined by:

$$f_{(x,y)} = \beta \min_{(s,t)\epsilon S_{(CP,x,y)}} p_{(s,t)} \qquad (4)$$

where, $S_{(CP,x,y)}$ and $\beta$ are the set of grid cells and technology-specific constant, respectively. Also, $p_{(s,t)}$ is the process variation information at grid point *(s,t)*.

*B. Reliability Models*

In this work, to provide a comprehensive reliability evaluation of the studied manycores, the different temperature-related aging mechanisms, including thermal cycling (TC), negative bias temperature instability (NBTI), and hot carrier injection (HCI), as well as the Electromigration (EM) phenomena are investigated. In the following, we detail each of these mechanisms.

TC occurs when the temperature undergoes a cycle of rising, dropping down, and then returning to its initial state [15]. As the number and amplitude of cycles increase, the cores aging increases significantly, i.e., the TC-related mean time to failure ($MTTF_{TC}$) is decreased. To quantify these cycles and calculate the $MTTF_{TC}$, Downing's simple rain flow-counting algorithm proposed in [18] is utilized. However, to determine the number of cycles ($N_{TC}$), the Coffin-Manson formula is employed, defined by [15]:

$$N_{TC}(i) = A_{TC}(\delta T_i - T_{th})^{-b} \exp\left(\frac{E_{aTc}}{KT_{max}}\right) \qquad (5)$$

where $A_{TC}$ and $b$ are empirical and |Coffin-Monson exponent constants, respectively. $T_{th}$ is the threshold temperature where the inelastic deformation starts, and $\delta T_i$ is the maximum thermal amplitude change of $i^{th}$ thermal cycle. Also, $E_{aTc}$ and $T_{max}$ are the activation energy and the maximum temperature during the cycle, respectively. Based on (5), the $MTTF_{TC}$ is calculated by [15]:

$$MTTF_{TC} = \frac{N_{TC}\sum_{i=0}^{m} t_i}{m} \qquad (6)$$

where $t_i$ represents the duration of each cycle and $m$ denotes the total number of cycles.

NBTI is a phenomenon that impacts the reliability of transistors in electronic devices. It occurs due to the gradual accumulation of charge traps within the gate oxide of a transistor when it experiences a negative bias voltage for an extended period. The MTTF variation caused by the NBTI phenomena is calculated by [19]:

$$MTTF_{NBTI} \propto \left(\left[\ln\left(\frac{A}{1 + 2e^{\frac{B}{KT}}}\right) - \ln\left(\frac{A}{1 + 2e^{\frac{B}{KT}}} - c\right)\right] \times \frac{T}{e^{\frac{-D}{KT}}}\right)^{\frac{1}{\beta}} \qquad (7)$$

where $A$, $B$, $C$, $D$, and $\beta$ are fitting parameters and $k$ is the Boltzmann constant.

HCI occurs when high-energy electrons, referred to as hot carriers, are generated within the channel region of MOSFETs due to the presence of a high electric field. These hot carriers can accumulate enough energy to penetrate the gate oxide, subsequently becoming trapped. This trapping leads to a modification of the oxide charge characteristics. The MTTF model of the HCI is defined by [20]:

$$MTTF_{HCI} \propto \left(\frac{I_{sub}}{W}\right)^{-n_{HCI}} \times e^{\frac{Q_{HCI}}{KT}} \qquad (8)$$

where $I_{sub}$ is the peak substrate current, $W$ is the width of the transistor, and $Q_{HCI}$ is the activation energy.

Finally, EM is caused by the migration of metal atoms triggered by momentum transfer between electrons and metal ions within the interconnects. This migration process can result in the creation of voids or hillocks within the interconnects, subsequently impacting the performance and reliability of the electronic system. The MTTF change of the EM mechanism is calculated by [20]:

$$MTTF_{EM} \propto I^{-n_{EM}} \times e^{\frac{Q_{EM}}{KT}} \qquad (9)$$

where $I$ is the current flowing into or out of the contact window, $n_{EM}$ is an empirical parameter with a value between 1 and 2, and $Q_{EM}$ is the electromigration activation energy.

*C. Reinforcement Learning Model*

The primary objective of Reinforcement Learning (RL) is to learn the optimal behavior within an environment to maximize the obtained reward. This target is achieved by iteratively interacting with a dynamic or uncertain environment, employing a trial-and-error approach. In this framework, the interaction between the learning agent and the environment is represented by a finite set of state space *S*, the available actions *A*, and a reward function *R: S×A → R* [7]. Through this process, the RL agent learns how to make effective decisions and take appropriate actions based on the observed state and associated rewards. RL algorithms can be categorized as either model-based or model-free. Model-free algorithms, unlike their model-based counterparts, do not construct an explicit model of the environment or the Markov Decision Process (MDP).

Table 1 Used notations.

| Notations used in formulas | |
| --- | --- |
| $W_{x,y}$ | Wire width |
| $H_{x,y}$ | Wire Height |
| $Res_{x,y}$ | Power grid resistance at the tile (x,y) |
| $\kappa_1, \kappa_2$, and $\gamma$ | Technology-specific constants |
| $f_{x,y}$ | Maximum frequency |
| $S_{CP,x,y}$ | Set of grid points |
| $\beta$ | Technology-specific constant |
| $N_{TC}$ | Number of cycles |
| $A_{TC}$ | Empirically constant |
| $b$ | Coffin-Monson exponent constant |
| $T_{th}$ | Threshold Temperature |
| $\delta T_i$ | Maximum thermal amplitude change in the $i^{th}$ thermal cycle. |
| $E_{aTc}$ | Activation energy |
| $T_{max}$ | Maximum temperature during the cycle |
| $t_i$ | Cycle duration |
| $m$ | Total number of cycles |
| $K$ | Boltzmann constant |
| $A, B, C, D,$ and $\beta$ | Fitting Parameters |
| $I_{sub}$ | Peak substrate current |
| $w$ | Width of transistor |
| $Q_{HCI}, Q_{EM}$ | Activation energy |
| $I$ | Current flowing into or out of the contact window. |

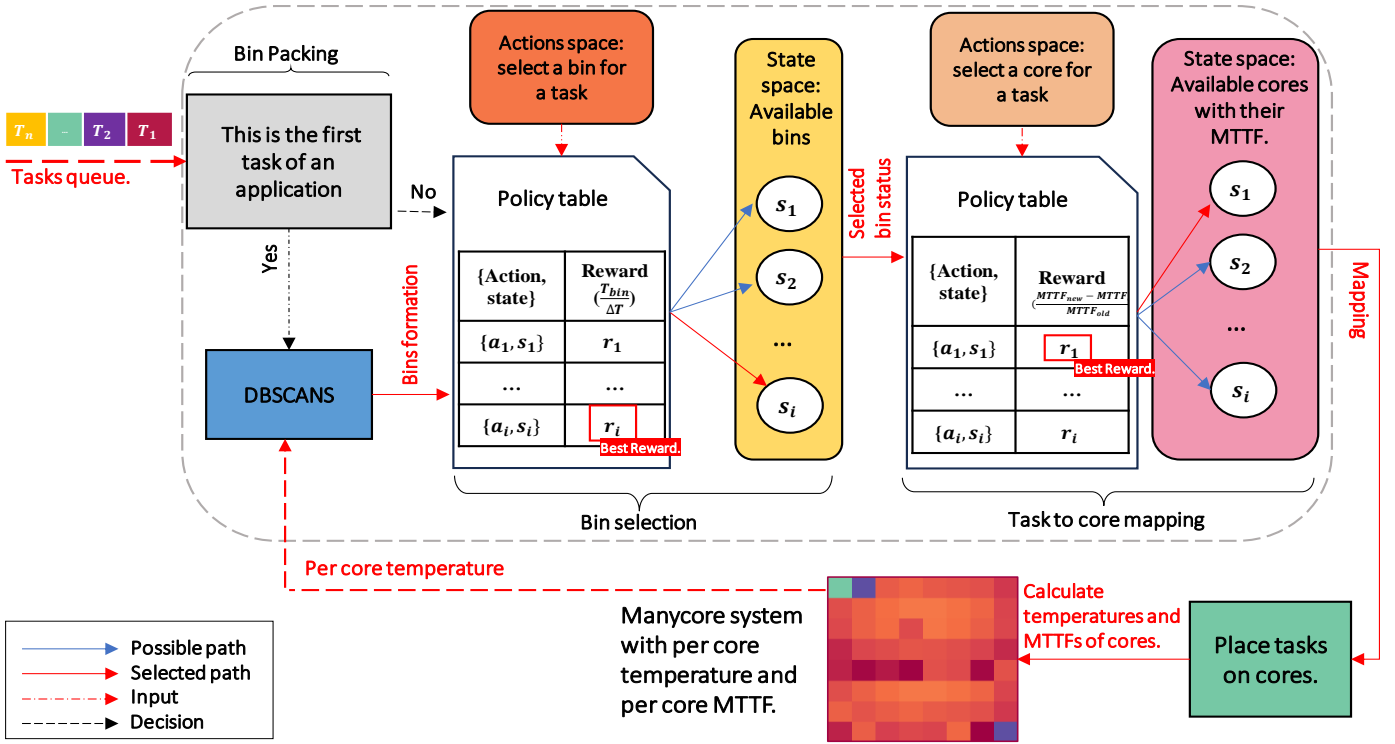

Figure 2 The flow of the proposed reinforcement learning (RL)-based task mapping method to improve the reliability of clustered manycores.

The dynamic nature of our environment and the complexities involved in task mapping for reliability justify the use of reinforcement learning. Our approach allows for real-time adaptability, efficient learning from experience, and optimized mappings that can significantly enhance system performance beyond what a well-designed heuristic could achieve.

### III. THE PROPOSED REINFORCEMENT LEARNING-BASED TWO-LEVEL TASK MAPPING

In this section, we introduce our proposed RL-based two-level task mapping technique, which mitigates the effects of aging mechanisms discussed in Subsection III.B, while considering the heat exchange between adjacent cores. This technique consists of three steps: DBSCAN-based bin packing, and task-to-bin and task-to-core mapping both using the Q-learning algorithm. A widely used model-free algorithm for RL is Q-learning. It estimates the quality of taking an action 'a' from a given state 's' by utilizing a Q-table that stores the Q-values for each state-action pair $(s,a)$. This value function, $Q(s,a)$, guides the agent in selecting the optimal action at a given state to maximize long-term rewards. Initially, the Q-values in the Q-table are assigned to random values. After receiving a reward, the Q-values are updated iteratively. The key principle of the Q-learning algorithm is the frequent updating of Q-values. The updates are performed by [7]:

$$Q'(s_k, a_k) = Q(s_k, a_k) +$$
$$\beta_k \cdot \left(r_{k+1} + \gamma + \max_{a \in A} Q(s_{k+1}, a) - Q(s_k, a_k)\right) \quad (10)$$

where $r_{(K+1)}$ is the expected reward at $t_{(K+1)}$ after taking action $a_K$ at $t_K$. $\gamma \in (0,1)$ is the discount factor and $\beta_K \in (0,1)$ is the learning rate at time $t_K$. Due to the high adaptability and flexibility of the Q-learning algorithm, as well as its ability to allow agents to learn without prior knowledge of the system, we employed this algorithm in our work; our tests show that cumulative rewards stabilize, and Q-values converge within 200 episodes in each configuration, supporting the model's adaptability and stability across different system scales. Figure 2 illustrates the flow of the proposed technique. As shown in this figure, cores are first packed into bins. Then, tasks are assigned to the bins and subsequently to the cores using reinforcement learning. In the following, we detail each step of the proposed mapping method.

*A. Bin Packing*

Inspired by [16], to consider the thermal effects of adjacent cores, we utilize a temperature-aware bin-packing approach. Towards this end, the cores are clustered (packed) into different bins based on their temperatures, using the density-based spatial clustering of applications with noise (DBSCAN) method [21]. DBSCAN is a density-based clustering algorithm that is capable of identifying clusters of various shapes and sizes within a large dataset, even in the presence of noise and outliers [22]. Using this algorithm, cores with temperatures close to each other are assigned to the same bins. Then, an arrived task can be assigned to a bin that performs that task resulting in minimum thermal variation among the cores inside the selected bin. Inputs of this step are the amount of maximum allowable difference between the temperatures of cores within a bin (*Epsilon*) determined empirically, and the parameter *M* that represents the minimum number of cores required to form a bin. Based on the explorations performed in [16], setting *Epsilon* to *0.7* and *M* to 1 result in maximum MTTF. Thus, we used these values in our experiments.

*B. Bin Selection*

In the bin selection step, a bin is selected to perform the arrived task such that the thermal variation among the cores/bins is minimized. Compared to the [16], in this work, the bin selection step is performed using the Q-learning algorithm, to remove the need for offline performing of application to extract temperatures of the tasks, required for the bin packing step. This step takes the set of formed bins, denoted as *{B_(b)*, $b \in \{1, ..., n_{Bin}\}\}$, as input, and outputs the task-bin pair *{task, selected bin}*. Note that the Q-learning algorithm relies on defining the state space, action space, and reward function, which are detailed in the following.

*1. State Space*

To achieve the target of the bin selection step (i.e., minimizing thermal variations among the cores and bins), we define the state space in the Q-learning algorithm, taking into account the temperatures of the bins represented by the average temperature of the cores inside each bin, and the status of each bin including the number of mapped tasks and the number of cores within each bin. By considering the temperatures of the bins, we can capture the thermal characteristics of each bin and their potential impact on the overall system aging. This information allows the Q-learning algorithm to make informed decisions regarding task-to-bin (bin selection) assignments. Additionally, incorporating the status of each bin provides a measure of the bin's capacity and workload. This information helps to ensure a balanced distribution of tasks among the bins, preventing overloading and promoting efficient utilization of resources.

*2. Action Space*

In this work, the action space is defined as the set of bins that the agent can choose from to assign a task. We denote the action space for task *i* as $A_i = \{a_{i,1}, ..., a_{i,n}\}$, where *i* represents the task number and *n* represents the number of bins. Initially, when the task-to-bin mapping process begins, there are no available actions. However, once the cores are packed (see Subsection III.A) and the number of bins and the temperature of each bin are determined, action space is created for tasks based on the current status of manycore system. It is notable that in each execution of the bin selection step, the action space is updated, i.e., the manycore system evolves and the available actions for tasks are dynamically adjusted. Based on this dynamicity that relies on the current state of the manycore system, the agent can be adapted and make informed decisions on the bin selection, considering the bin formation and temperature distribution. This flexibility offers efficient task allocation and assists in reducing thermal variations within the manycore system. While reinforcement learning (RL) is typically implemented with static action and state spaces, recent studies have explored frameworks incorporating dynamic action spaces to better address real-time and evolving environments. For instance, in [23], the authors examine the challenges associated with dynamic action spaces in non-stationary MDPs and propose an intelligent Action Pick-up (AP) algorithm. This algorithm enables agents to autonomously select valuable actions from new and unseen options, thereby enhancing learning efficiency. Inspired by this approach, we adapt the action space in our model to effectively respond to real-time temperature fluctuations across cores.

*3. Reward*

The reward function is defined based on the state and the action taken by the RL agent. Since the actions leading to lower thermal variation are more desirable, the reward function should incorporate the temperature of each bin and the temperature variation observed in the selected bin. Thus, the reward function is defined by:

$$r(s_t, a_{t,n}) = \frac{T_{bin}}{\Delta T} \qquad (11)$$

where $s_t$, $a_{t,n}$, and $T_{bin}$ are the state of the environment, the performed action, and the temperature of the selected bin, respectively. Note that *r* is employed to update the *Q-values* as per (10), which then used to construct the *Q-table*.

*C. Task to Core Mapping*

After the bin selection, in this step, the arrived task is mapped to a core within the selected bin. Note that compared to the [16], the task to core mapping step is also performed using the Q-learning algorithm. In the following, we define the state space, action space, and reward function for this step.

*1. State Space*

The task-to-core mapping step is performed regarding the reliability of each core. In this work, the MTTF parameter is considered as the reliability of cores, which indicates the average lifespan of a core before it fails. Therefore, the state space is defined based on the average MTTFs (the four different MTTFs discussed in Subsection III.B) of cores and their current status, i.e., whether they are free or busy.

*2. Action Space*

For this step (i.e., task to core mapping), the action space is defined as the choices for selecting a core within the bin to perform the arrived task from the execution queue. Note that the number of cores within each bin may vary, leading to different action space sizes for tasks in different bins. For example, if bin *B1* has three cores and bin *B2* has four cores, the action space for tasks in *B1* consists of three choices, while the action space for tasks in *B2* is composed of four choices. Thus, the action spaces should be dynamically adjusted based on the number of cores within each bin.

*3. Reward*

The reward function needs to incorporate the MTTF of each core and the MTTF difference that occurs after executing a task. Since the objective is to maximize the MTTF of each core, a higher reward should be assigned to a greater MTTF difference. Thus, the reward function for this step is defined as:

$$r(s_t, a_{t,n}) = \frac{MTTF_{new} - MTTF_{old}}{MTTF_{old}} \qquad (12)$$

where $s_t$ is environment state, $a_{t,n}$ is the action agent takes, $MTTF_{new}$ is MTTF of the core after task execution, and $MTTF_{old}$ is MTTF of the core before the task execution. Note that the MTTF values in (12) are the average of equations (6) to (9). In the final step, the calculated $r$ is utilized to update the $Q$-values based on equation (10).

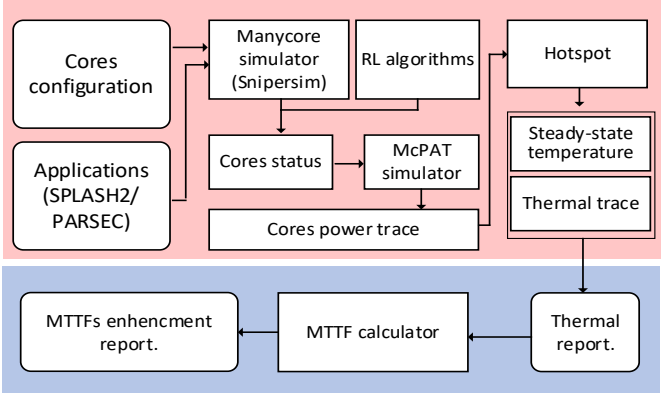

Figure 3. Experimental flow and simulation setup.

## IV. RESULTS AND DISCUSSIONS

In this section, first, we detail the experimental setup developed and used in our examinations. Then, we assess the effectiveness of our proposed RL-based mapping method by studying its reliability for different manycore systems under the various application benchmark suites. We also compared the results of our proposed RL-based mapping method with the state-of-the-art works.

### A. Experimental setup

The experimental flow and simulation setup of our examinations are illustrated in Figure 3. As shown in this figure, cores configuration and applications of benchmarks are fed as the input to the *Snipersim* simulator [24], to execute the applications. The Hotspot simulator is also employed to obtain the temperature profile of the examined manycores [25].

Note that the evaluations were conducted on manycores implemented in a 22-nm technology, with tiles sized at 0.7 mm × 0.8 mm. Each tile in the system consists of a Nehalem core with a private L1 cache (256 kB) and a private L2 cache (512 kB). In our experiments, we executed different set of workloads comprising fifteen applications from the SPLASH2 [26] and PARSEC [27] benchmark suites, running on 16, 32, and 64 manycore systems. Moreover, the values of parameters used in equations (5) to (10) to perform the simulations were shown in Table 2.

### B. Reliability Evaluation

In our evaluations, we compare our proposed technique with the random task mapping and the conventional TC-based task mapping, as well as the state-of-the-art techniques proposed in [15] and [16] for reliability improvement of manycores considering the TC, EM, HCI, and NBTI aging mechanisms.

Table 3 presents the significant findings of our study conducted on 16, 32, and 64 cores systems, showcasing noteworthy improvements in $MTTF_{TC}$, $MTTF_{EM}$, $MTTF_{HCI}$, and $MTTF_{NBTI}$.

In the 16 cores system, compared to the four other techniques, we achieved an average improvement of 24% in $MTTF_{TC}$, along with enhancements of 6% for $MTTF_{EM}$, 12% for $MTTF_{HCI}$, and 7% for $MTTF_{NBTI}$. For the 32 cores system, our proposed approach demonstrated on average 36%, 20%, 12%, and 10% improvement in $MTTF_{TC}$, $MTTF_{EM}$, $MTTF_{HCI}$, and $MTTF_{NBTI}$, respectively. Moreover, in the case of the 64 cores system, notable average enhancements were observed, where our method achieved an improvement of 27% in $MTTF_{TC}$, 19% in $MTTF_{EM}$, 6% in $MTTF_{HCI}$, and 8% in $MTTF_{NBTI}$. In general, for the four studied aging mechanism, our proposed mapping technique achieved on average, 28%, 17%, 11%, and 7% MTTF improvement, compared to the random, TC-based, [15], and [16] techniques, respectively.

Figure 4 illustrates heatmaps of a 64 cores system after executing 9 applications from SPLASH2 and 6 applications from PARSEC benchmarks, employing the aforementioned task mapping techniques. The figure clearly shows that our proposed technique results in a significantly more homogeneous thermal map compared to other methods. This improved thermal management leads to a lower average core temperature, directly enhancing the system's MTTF, as quantified below each part of the figure.

Table 2 Values of Parameters of equations (5) to (10)

| Parameter | Value | Used in |
|---|---|---|
| $E_{aTc}$ | 0.42 eV | Eq. (5) |
| $b$ | 2.35 | Eq. (5) |
| $T_{th}$ | 1°C | Eq. (5) |
| $K$ | 8.62E-05 eV/K | Eq. (5) |
| $Q_{HCI}$ | 0.25 | Eq. (8) |
| $n_{HCI}$ | 3 | Eq. (8) |
| $Q_{EM}$ | 0.9 eV | Eq. (9) |
| $n_{EM}$ | 1.1 | Eq. (9) |
| $\gamma$ | 0.28 | Eq. (10) |
| $\beta_K$ | 0.72 | Eq. (10) |

Table 3 The average MTTF improvements of our proposed method vs. state-of-the-art techniques, when performing fifteen applications from the SPLASH2 and PARSEC benchmark suites, running on 16, 32, and 64 manycore systems.

| MTTF | Number of cores | State-of-the-art task mapping techniques | | | |
|---|---|---|---|---|---|
| | | Random | TC-based | [15] | [16] |
| **$MTTF_{TC}$** | 16 | 33% | 27% | 26% | 10% |
| | 32 | 66% | 34% | 27% | 17% |
| | 64 | 49% | 29% | 22% | 7% |
| **$MTTF_{HCI}$** | 16 | 25% | 13% | 5% | 6% |
| | 32 | 34% | 22% | 15% | 10% |
| | 64 | 32% | 21% | 14% | 8% |
| **$MTTF_{EM}$** | 16 | 13% | 8% | 2% | 2% |
| | 32 | 25% | 13% | 6% | 4% |
| | 64 | 14% | 9% | 1.8% | 1.3% |
| **$MTTF_{NBTI}$** | 16 | 12% | 11% | 4% | 3% |
| | 32 | 16% | 13% | 6% | 5% |
| | 64 | 12% | 9% | 5.1% | 5.5% |

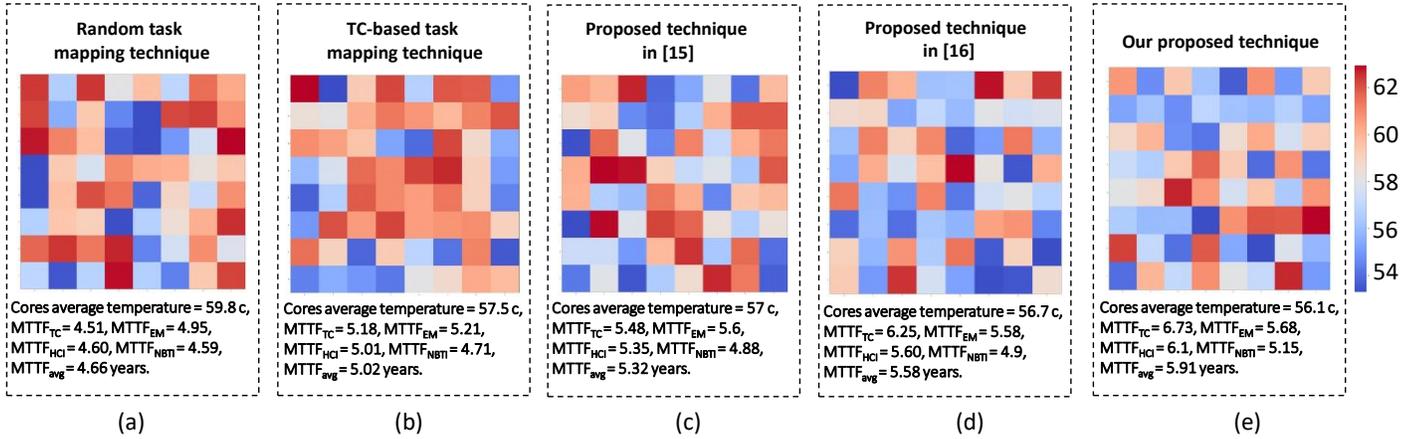

Figure 4 Heatmaps of a 64 cores system after executing 9 applications from SPLASH2 and 6 applications from PARSEC benchmark suites for a) Random, b) TC-based, c) [15], d) [16], and e) our proposed task mapping techniques.

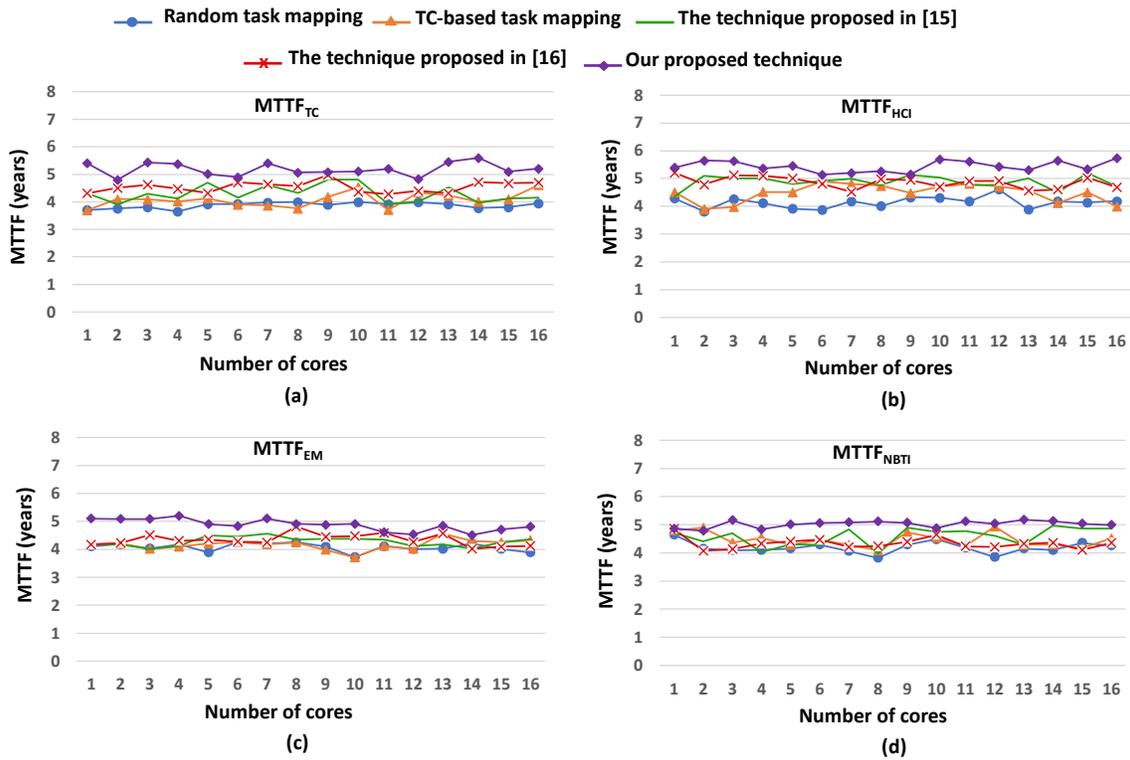

Figure 5 Average MTTF of a 16 cores system under the a) TC, b) HCI, c) EM, and d) NBTI aging mechanism, when performing 9 applications from SPLASH2 and 6 applications from PARSEC benchmark suites.

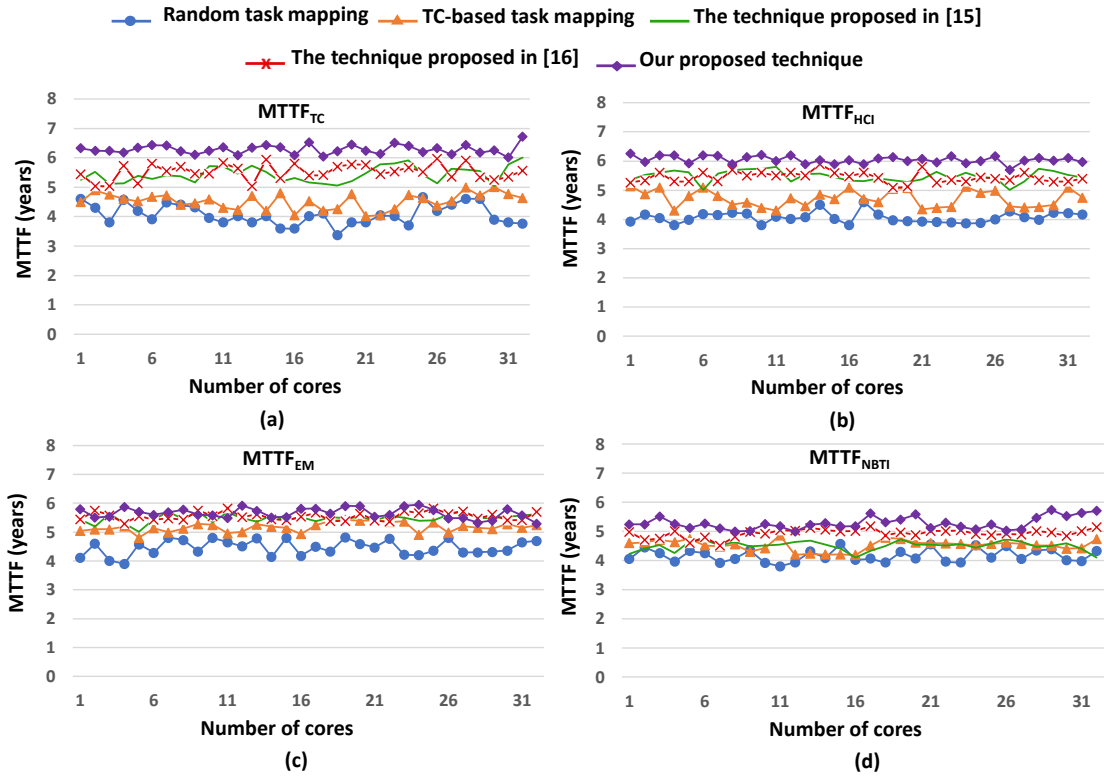

Figure 6 Average MTTF for a 32 cores system using a) TC, b) HCI, c) EM, and d) NBTI aging mechanism, when performing 9 applications from SPLASH2 and 6 applications from PARSEC benchmark suites.

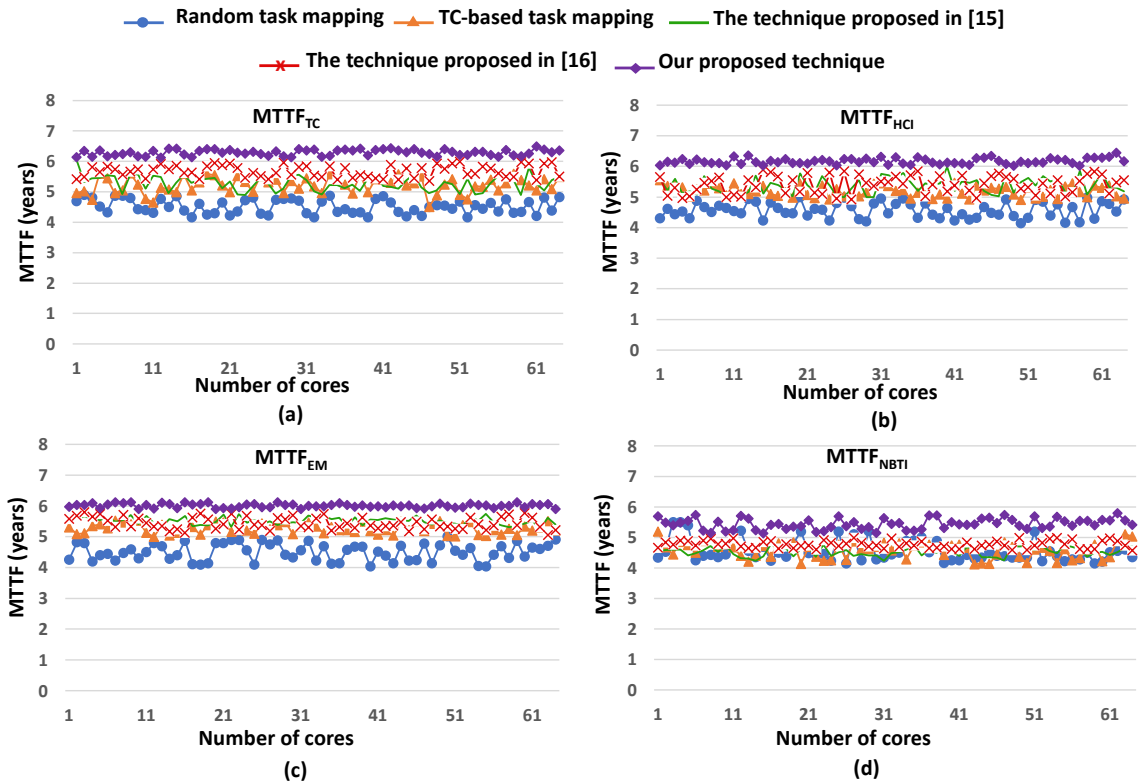

Figure 7 Average MTTF for a 64 cores system using a) TC, b) HCI, c) EM, and d) NBTI aging mechanism, when performing 9 applications from SPLASH2 and 6 applications from PARSEC benchmark suites.

Figure 5 illustrates the MTTF of the cores within a 16 cores system, when performing 9 applications from SPLASH2 (FFT, LU, RADIX, Cholesky, FMM, Ocean, Barnes, Raytrace, and Radiosity) and 6 applications from PARSEC (Blackscholes, Canneal, Dedup, X264, Vips, and Swaptions) benchmark suites. Based on the results, upon employing our proposed RL-based technique, the average value of $MTTF_{TC}$, $MTTF_{HCI}$, $MTTF_{EM}$, and $MTTF_{NBTI}$ are 5.19, 5.37, 4.73, and 4.93 years, respectively. Comparatively, for the random mapping technique, TC-based technique, the approach described in [15], and the method presented in [16], the corresponding four tuple MTTFs are (3.88, 4.29, 4.17, and 4.38), (4.07, 4.73, 4.36, and 4.44), (4.7, 5.03, 4.63, and 4.75), and (4.11, 5.10, 4.60, and 4.71) years, respectively. Thus, compared to the all studied mapping methods, our proposed RL-based mapping method achieved on average (24%, 12.25%, 6%, and 7%) improvements in the four $MTTF_{TC}$, $MTTF_{HCI}$, $MTTF_{EM}$, and $MTTF_{NBTI}$ aging mechanisms, respectively.

Figure 6 shows the MTTF of the cores in a 32 cores system, comparing state-of-the-art works with our RL-based proposed technique across various aging mechanisms. Based on the results, our proposed technique demonstrates an average $MTTF_{TC}$ of 6.24 years, $MTTF_{HCI}$ of 5.81 years, $MTTF_{EM}$ of 5.66 years, and $MTTF_{NBTI}$ of 5.63 years within a 32 cores system. In this case, compared to the other studied mapping methods, our one achieved on average 36%, 20.25%, 12%, and 10% increments in the $MTTF_{TC}$, $MTTF_{HCI}$, $MTTF_{EM}$, and $MTTF_{NBTI}$, respectively.

Finally, Figure 7 illustrates the MTTF of cores within a 64 cores system, where upon implementing our technique in this system, the average $MTTF_{TC}$, $MTTF_{HCI}$, $MTTF_{EM}$, $MTTF_{NBTI}$ are 6.73, 6.10, 5.68, and 5.15 years, respectively. These findings demonstrate an improvement of up to 49% when compared to the MTTFs achieved by the four previously discussed techniques.

## V. Conclusion

This paper presents a Reinforcement Learning (RL)-based task mapping technique to improve the reliability of manycore systems, while mitigating the NBTI, HCI, TC, and EM effects, simultaneously. In the proposed technique, at first, using the DBSCAN method, the cores are clustered into different bins based on their temperature profiles. Then, leveraging the Q-learning algorithm, an appropriate bin that minimizes thermal variations among the cores/clusters is selected for performing the arrived task. Subsequently, the task is mapped into a given core inside the selected cluster that ensures the system reliability considering the process variation information, which may affect the voltage and frequency of the cores. Based on the results, a substantial improvement in average MTTF up to 25% has been achieved compared to the state-of-the-art works.

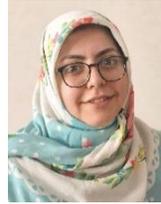

**Fatemeh Hossein-Khani** received her B.Sc. degree from Amirkabir University of technology, Tehran, Iran, and M.Sc. degree from the Tarbiat Modares University, Tehran, Iran, both in Computer Engineering, in 2017 and 2022, respectively. Her research interests include fault-tolerant system design, reconfigurable systems, AI and machine learning hardware and system-level design, and hardware accelerators.

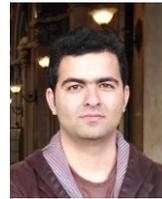

**Omid Akbari** received the B.Sc. degree from the University of Guilan, Rasht, Iran, in 2011, the M.Sc. degree from Iran University of Science and Technology, Tehran, Iran, in 2013, and the Ph.D. degree from the University of Tehran, Iran, in 2018, all in Electrical Engineering, Electronics - Digital Systems sub-discipline. He was a visiting researcher in the CARE-Tech Lab. at Vienna University of Technology (TU Wien), Austria, from Apr. to Oct. 2017, and a visiting research fellow under the Future Talent Guest Stay program at Technische Universität Darmstadt (TU Darmstadt), Germany, from Jul. to Sep. 2022. He is currently an assistant professor of Electrical and Computer Engineering at Tarbiat Modares University, Tehran, Iran, where he is also the Director of the Computer Architecture and Dependable Systems Laboratory (CADS-Lab). His current research interests include embedded machine learning, reconfigurable computing, energy-efficient computing, distributed learning, and fault-tolerant system design.